\documentclass[journal]{IEEEtran}
\usepackage{graphicx}
\usepackage{epstopdf}
\usepackage{cite}
\usepackage{makecell}
\usepackage{array}
\usepackage{multirow}
\usepackage{amsmath}
\usepackage{amssymb}

\begin{document}
\newcolumntype{R}[1]{>{\raggedleft\arraybackslash}p{#1}}
\newcolumntype{L}[1]{>{\raggedright\arraybackslash}p{#1}}
\newcolumntype{C}[1]{>{\centering\arraybackslash}p{#1}}

%
\title{Band Attention Convolutional Networks For Hyperspectral Image Classification}

\author{Hongwei~Dong,
        Lamei~Zhang,~\IEEEmembership{Member,~IEEE,}
        Bin~Zou,~\IEEEmembership{Senior Member,~IEEE}
\thanks{This work was supported in part by the National Natural Science Foundation of China under Grant 61401124 and 61871158, in part by Scientific Research Foundation for the Returned Overseas Scholars of Heilongjiang Province under Grant LC2018029.}
\thanks{The authors are with the Department of Information Engineering, Harbin Institute of Technology, Harbin, 150001, China (e-mail :lmzhang@hit.edu.cn).}}



\markboth{}
{Shell \MakeLowercase{\textit{et al.}}: Bare Demo of IEEEtran.cls for IEEE Journals}

\maketitle

\begin{abstract}
Redundancy and noise exist in the bands of hyperspectral images (HSIs). Thus, it is a good property to be able to select suitable parts from hundreds of input bands for HSIs classification methods. In this letter, a band attention module (BAM) is proposed to implement the deep learning based HSIs classification with the capacity of band selection or weighting. The proposed BAM can be seen as a plug-and-play complementary component of the existing classification networks which fully considers the adverse effects caused by the redundancy of the bands when using convolutional neural networks (CNNs) for HSIs classification. Unlike most of deep learning methods used in HSIs, the band attention module which is customized according to the characteristics of hyperspectral images is embedded in the ordinary CNNs for better performance. At the same time, unlike classical band selection or weighting methods, the proposed method achieves the end-to-end training instead of the separated stages. Experiments are carried out on two HSI benchmark datasets. Compared to some classical and advanced deep learning methods, numerical simulations under different evaluation criteria show that the proposed method have good performance. Last but not least, some advanced CNNs are combined with the proposed BAM for better performance.
\end{abstract}


\IEEEpeerreviewmaketitle

\section{Introduction}
\IEEEPARstart{h} {yperspectral} images (HSIs) contain hundreds of near-continuous spectral bands and this benefits not only attracts the attention in the field of remote sensing, but also arouses great interest in some other fields. With the development of hyperspectral sensors, the intelligent interpretation of HSIs has become an important research issue in the field of remote sensing. However, this problem has not been well solved due to many factors. One important reason for this is that the high-dimensional data of HSIs contains a considerable degree of noise and redundancy, although some traditional band selection methods based on prior knowledge or data characteristics have been extensively studied \cite{Su2018A,Wei2018Matrix}.
\par The deep learning technique, represented by convolutional neural networks (CNNs) \cite{Lecun2014Backpropagation}, has attracted extensive attention due to its good performance in recent years. As a continuation of statistical machine learning \cite{Vapnik1995}, deep learning has powerful ability of data fitting under sufficient supervisory information. As this technique drives to maturity, some CNNs based methods \cite{vgg,ResNet,DenseNet} have been able to match the accuracy of human recognition in some specific visual tasks.
\par Naturally, we think about using the capability of CNNs to solve the problem of HSI classification. In \cite{Chen2014Deep}, deep learning method was firstly applied to HSIs for classification. The two-stream and 3D networks \cite{twostream,C3D} used for video processing in computer vision has been widely used in HSIs classification \cite{Yang2016}. Because HSIs contain spectral dimension and spatial dimension \cite{Li2018}, which is similar to the relationship between spatial dimension and temporal dimension in videos. Although some fairly advanced algorithms have been applied to HSI classification, there are still two major problems of deep learning in HSIs. One is the lack of weakly supervised algorithms, which is caused by the difficulty in obtaining data and labels of HSIs. The other, also studied in this letter, is that the network models are mostly borrowed from the mainstream networks in RGB image processing rather than customized for HSIs. The connotation of deep learning lies in that: models are constructed to adapt to the input data in different types, and the deep representations can be obtained through the model so as to get accurate generalization performance. Therefore, it is necessary to design a network model which is highly compatible with the characteristics of HSIs.
\par Based on the above analysis, the purpose of this letter is to construct a classification network for HSIs, which can adapt to the problem of band redundancy. Inspired by the attention mechanism \cite{attention1,attention2,SENET}, we propose a band attention based HSIs classification framework in this letter. In detail, a band attention module (BAM) embedded in the classification network is proposed. The proposed BAM obtains the global information through a series of convolutions and generates the required weight vector for the processing of input bands. It aims to be a plug-and-play complementary component of the existing HSIs classification networks and also orthogonal and complementary to methods that focus on spatial attention \cite{chen2019}. For the input data, the proposed model firstly carries out band selection (or weighting, which is determined by the last activation before the weight vector is obtained) through BAM, then obtains the recognition results through the classification module. The whole network is training end-to-end and the processing of bands shares supervisory information with image classification. The validity of the proposed method is demonstrated by the numerical experiments on HSIs benchmark datasets.
\par The rest of this letter is organized as follows: The proposed strategies are listed in Sections \uppercase\expandafter{\romannumeral2}. Experimental results are exhibited in Section \uppercase\expandafter{\romannumeral3}. Conclusion and future directions are given in Section \uppercase\expandafter{\romannumeral4}.

\section{Proposed Method}
\label{sec:2}
Fig. 1 shows the flowchart of the proposed band attention convolutional neural network (BACNN).
\begin{figure}[h]
\begin{centering}
\includegraphics[width=8.5cm]{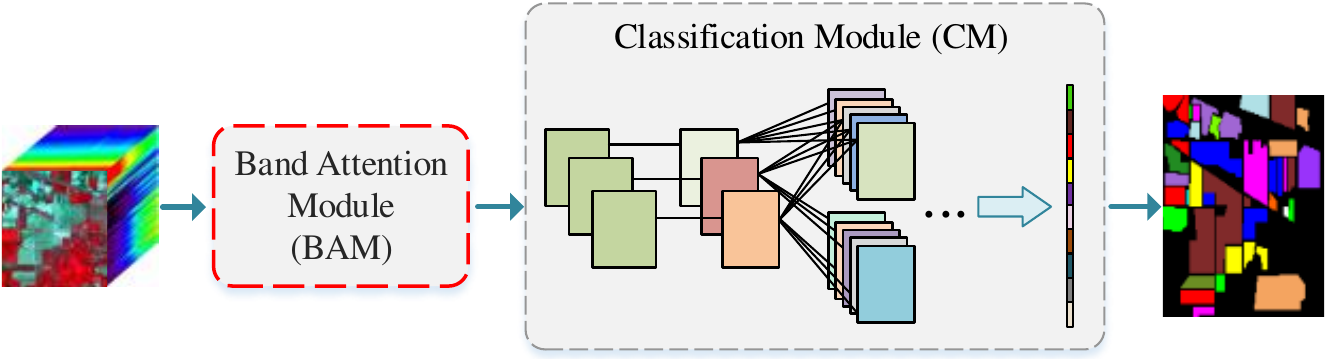}
\caption{General flow chart of BACNN. The part of red dotted line is unique to the proposed method compared with the ordinary CNNs.}\label{fig1}
\end{centering}
\end{figure}
\par As shown in Fig. 1, compared with the ordinary CNN, the proposed BACNN has a BAM which can selectively select the input bands. In this way, the adverse effects of redundancy and noise of the bands on CNN classification are reduced and the accuracy is improved. Specifically, a $c\times 1$ weight is obtained by the BAM for a $h\times w\times c$ (c is the number of input bands) HSI and the input is channel-wise multiplied with the weight to obtain the band-processed HSI. Then we use a CNNs based network to classify the band-processed HSI. We think that this end-to-end model has better adaptive ability than the phased classification methods.
\subsection{Band attention module}
Attention mechanism has been widely used in image processing \cite{attention2} since it can adaptively stimulate or suppress the input information. Its core lies in infusing global information into the algorithm through the learning of an image mask, so as to accelerate the areas which are beneficial to improving accuracy. We use the attention mechanism as a tool for choosing wanted bands. The structure of BAM is depicted in Fig. 2.
\begin{figure}[h]
\begin{centering}
\includegraphics[width=8.0cm]{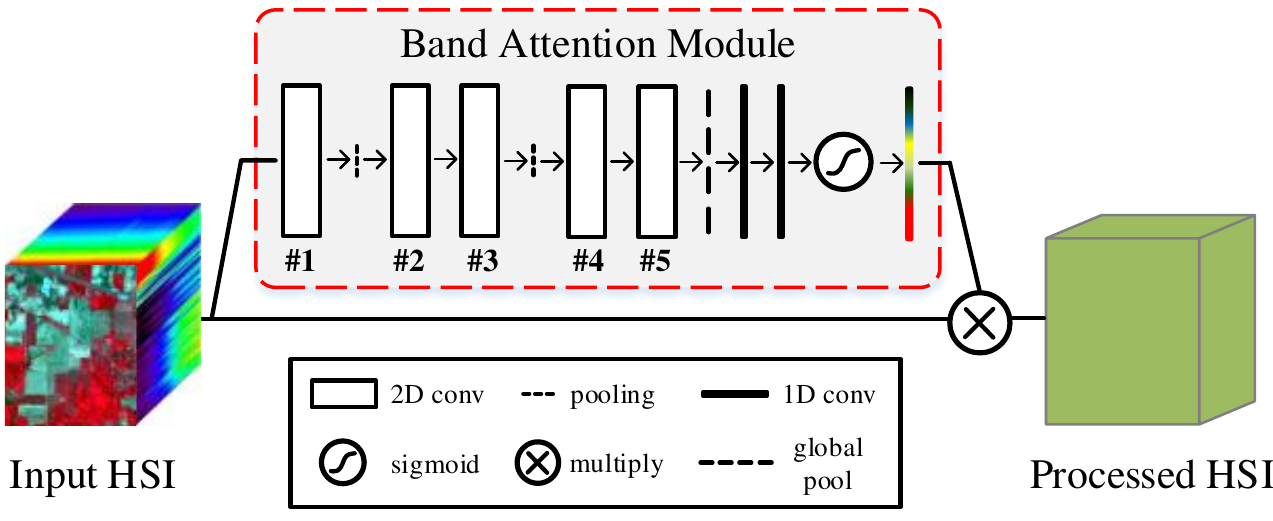}
\caption{The structure of the proposed BAM for attention based band processing.}\label{fig2}
\end{centering}
\end{figure}
\par As shown in Fig. 2, we use a series of convolutions and sub-sampling. The aim of these operations is to obtain the global information by reducing the resolution while expanding the receptive field and finally obtain the required weight vector. The used BAM consists of five $3\times 3$ 2D convolution layers which can be divided into three stages by two pooling layers, and each stage with the depth of 16, 32 and 32. Then two 1D convolution layers are used for further nonlinear learning between channels. Here has a hyperparameter $r$ to control the degree of information aggregation in 1D convolution layers. The first 1D kernel should be of size $1\times1\times 32\times c/r$ and the second with $1\times1\times c/r\times c$. Thus, a $c$ dimensional weight vector with global information of bands is learned, which can be seen as the mask of bands. Then the band mask is applied to the input HSI and the band-processed image can be obtained. The forward propagation of BAM can be expressed as:
\begin{equation}
H_{out}=\sigma_2(W_{22}\sigma_1(W_{21}f_{globalpool}(W_1H_{in})))\label{1}
\end{equation}
where $H_{in},H_{out}$ denote the input HSI and the output of BAM, $W_1,W_2$ represent 2D kernel matrix and 1D kernel matrix, $f_{globalpool}(\cdot)$ is used to fully fuse the spatial information contained in the feature maps and provides the basis for forming the band masks, which can be defined as:
\begin{equation}
f_{globalpool}(x_z)=\frac{1}{h' \times w'}\sum_{i=1}^{h'} \sum_{j=1}^{w'} x_z(i,j)\label{2}
\end{equation}
where $x_z$ means the $z$th feature map and $h',w'$ denote its current hight and width. $\sigma_1$ is rectified linear unit (ReLu) \cite{ReLU}. Maybe not the optimal, $\sigma_2$ is set to be sigmoid activation because this option performs better than other alternatives in the experiments.
\par This design is similar to the ``squeeze-and-excitation block'' in SENet \cite{SENET}. The difference is that we use convolutions instead of the SE block's global pooling to reduce the spatial resolution of the input. The reason for this change is that the BAM and SE block act on different objects. For an image, the value of each pixel represents only physical meaning, but not the feature. So using SE block with global pooling for band selection can not inject spatial global information. Besides, the goal of SE block is to adaptively select useful features while suppressing less useful ones in channel dimension of the feature maps so SE block is embedded in every convolution layer in SENet. Our aim is to process the input bands in order to reduce the side-effects of noise and redundancy in bands on classification. Therefore, the BAM only appears once in the proposed BACNN.
\par In fact, the design of BAM implies the idea of re-weighting \cite{IRLS} in statistical robust learning. Visual attention mechanism is essentially similar to the idea of re-weighting, the former is currently used in some deep learning models \cite{attention1}, while the latter is mostly used in shallow learning models. This also proves that the BAM embedded classification model has better robustness to the noise of the input bands.
\par Depth fixed vanilla convolution is considered to observe the effectiveness of the BAM in this letter, while various depth settings and advanced convolution operations such as dilated \cite{Yudilated}, 3D \cite{C3D} or depthwise separable \cite{xception} convolution also can be considered for better classification performance.
\subsection{BAM based HSI Classification}
With the former module for band selection, a classification module (CM) is also essential to classify the processed HSIs. It is worth to note that although the description is separate, BAM and CM are in the same network and training end-to-end. This integration of separated components can share supervisory information and has better generalization performance. In this letter, a basic CM based on VGGNet \cite{vgg} is mainly used, which can be seen from Fig. 3.
\begin{figure}[h]
\begin{centering}
\includegraphics[width=8.0cm]{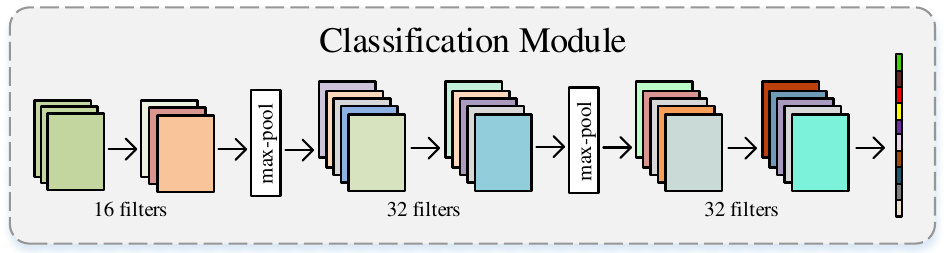}
\caption{The structure of the CM for classifying the band-processed HSIs.}\label{fig3}
\end{centering}
\end{figure}
\par It can be seen from Fig. 3 that a eight layers VGGNet is chosen as the CM. In most of the experiments in this letter, we use such a fairly simple classification network to classify the band-processed HSIs. The reason for this is that our aim is to infuse the capability of band selection into existing classification networks, rather than to study a more advanced one. A simple designed CM is enough to observe the improvement of accuracy after adding the proposed BAM. Nevertheless, some advanced CNNs also can be involved to achieve better classification performance, including two-stream CNN \cite{twostream,Li2018,Yang2016}, ResNet \cite{ResNet,8445697}, DenseNet \cite{DenseNet} and so on.
\subsection{Implementation details}
For the classical methods which require lots of engineering by hand, band selection or band weighting are two different stories. In the proposed method, we can selectively do band selection or band weighting by changing the last activation function in the BAM. There are several options: ReLu activation to achieve the weights with the value of zero or one for band selection; Sigmoid or softmax activation to let the weights in $[0,1]$ for band weighting.
\par In order to accelerate the convergence of training, we followed some mainstream designs: Before each convolution layer, we add a batch normalization layer \cite{Ioffe2015Batch} and a ReLU layer. Adam optimization method \cite{adam} with the learning rate of 0.0001 is chosen to achieve good training. Besides, $20\%$ neurons of fully connected layers in CM are randomly discarded to prevent overfitting in the training stage.
\section{Experiments}\label{sec:3}
\subsection{Datasets}
Numerical experiments are carried out on two benchmark HSI datasets including Indian Pines dataset and Kennedy Space Center (KSC) dataset to evaluate the effectiveness of the proposed method. Several state-of-the-art alternatives are chosen for comparison. The experiment environment: PC with Intel i7-7700 CPU, Nvidia GTX-1060 GPU (6 GB memory), and 16 GB RAM. Overall accuracy (OA), average accuracy (AA), and kappa coefficient are chosen as criteria to evaluate the performance in our experiments.
\par The Indian Pines dataset was acquired by the Airborne Visible/Infrared Imaging Spectrometer (AVIRIS) sensor over the Indian Pines test site in North-western Indiana and contains 200 bands after removing water absorption bands. The image consists of $145\times145$ pixels and there are 16 classes of land covers. 10249 pixels are selected for manual labeling according to the ground truth map. For the Indian Pines dataset, the number of labeled samples varies greatly among different classes. The smallest class ``Oats'' with only 20 labeled pixels, and the largest class ``Soybean-mintill'' with 2455 labeled pixels. The imbalance between categories undoubtedly brings difficulties for subsequent processing. Thus, $30\%$ samples of the classes with fewer samples and 80 samples of the richer classes are randomly chosen as training set, the remaining as testing set. In addition, for the classes with fewer samples, replication operations have been carried out to mitigate the negative impact of imbalanced classification. The false-color composite image and the corresponding ground reference map are demonstrated in Fig. 4.
\par The KSC dataset was acquired by the NASA AVIRIS sensor over the Kennedy Space Center, Florida. The HSI contains 176 bands after removing water absorption and low SNR bands. $512\times614$ pixels and 13 classes of land covers exist in the image. The number of labeled samples in this dataset is roughly same among different classes and 5211 pixels are selected for manual labeling according to the ground truth map. $10\%$ of the total are randomly chosen as training set and the remaining as testing set. The false-color composite image and the corresponding ground reference map are demonstrated in Fig. 5.
\begin{figure}[h]
\begin{centering}
\includegraphics[width=9.0cm,height=5.0cm]{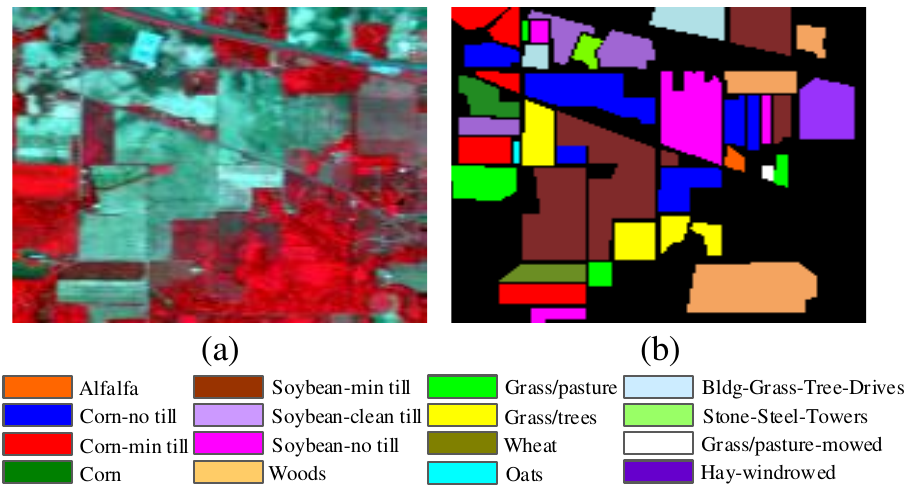}
\caption{Indian Pines dataset. (a) Three-channel false-color composition (bands 17, 27, and 50 for RGB). (b) Ground truth map.}\label{fig3.1}
\end{centering}
\end{figure}
\begin{figure}[h]
\begin{centering}
\includegraphics[width=9.0cm,height=5.0cm]{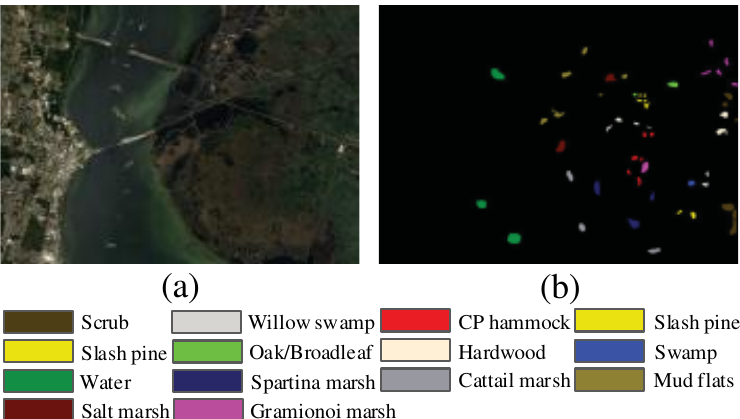}
\caption{KSC dataset. (a) Three-channel false-color composition (bands 10, 19, and 28 for RGB). (b) Ground truth map.}\label{fig3.2}
\end{centering}
\end{figure}

\subsection{Analysis of experimental results}
Table \uppercase\expandafter{\romannumeral1} shows the classification results of the proposed BACNN and other classification methods on the Indian Pines dataset. The size of the image slice is set to $15\times 15$ during the experiments. In order to test the effect of BAM on classification, we used an eight layers VGGNet as the CM in the experiments. The CM in the table means directly use VGGNet for classification and not to do any processing for input bands. The other three methods use different ways to select or weight the input bands, including SE block \cite{SENET} (SE+CM), band weighting module \cite{BandWeighting} (BW+CM) and the proposed BAM (BAM+CM). Although the band weighting module (BW) also uses the attention mechanism to adapt to HSIs, it only acts on the spectral dimension and ignores the abundant information in the spatial dimension. In order to maintain the persuasiveness of the experiments, we keep a same depth and conduct ten times experiments for all involved models to eliminate randomness.
\par From Table \uppercase\expandafter{\romannumeral1} we know that the proposed BAM achieves the optimal classification performance on Indian Pines dataset. The 1D-BW is not enough to improve the classification results, but its accuracy is inferior to that of 2D no band processing network. Further, since the SE block is designed for feature but not for bands, it does not improve the classification accuracy of CM. What's worse is that it degrades the CM by $1\%$ of OA and $1.2\%$ of Kappa. The performance of these two shows that an improper band selection module not only fails to improve the performance of CM, but also has the opposite effect. In contrast, the accuracy of CM has been improved to a certain extent after adding BAM since the BAM is not only tailored for HSI, but also considers spatial and spectral information. The $2\%$-$3\%$ improvement of each criterion confirms the validity of the proposed BAM.
\begin{table}[!h]
\centering  
\caption{Classification Results on Indian Pines Dataset. Numbers in the Parenthesis is the Standard Variances of the Accuracies Obtained in Repeated Experiments.}
\renewcommand\arraystretch{1.25}
\begin{tabular}{C{0.5cm}C{1.5cm}C{1.55cm}C{1.62cm}C{1.5cm}}  
\hline
Class    &CM &SE+CM \cite{SENET} &BW+CM \cite{BandWeighting} &BAM+CM \\
\hline
1     &97.88(2.05) &\textbf{98.18(2.12)} &90.30(9.98) &97.88(2.05)\\         
2     &85.89(2.62) &83.46(3.44) &55.70(10.28) &\textbf{89.82(1.82)}\\         
3     &92.65(3.12) &90.32(3.09) &72.50(9.71) &\textbf{95.68(1.47)}\\         
4     &\textbf{98.68(1.16)} &98.26(1.28) &89.65(4.98) &98.54(1.20)\\         
5     &93.63(1.90) &93.29(2.40) &79.34(6.06) &\textbf{94.92(1.61)}\\         
6     &97.18(1.67) &97.07(1.13) &80.67(9.97) &\textbf{98.27(0.86)}\\         
7     &99.50(1.58) &\textbf{100.00(0.00)} &92.5(10.07) &\textbf{100.00(0.00)}\\         
8     &99.97(0.09) &\textbf{100.00(0.00)} &96.24(3.81) &\textbf{100.00(0.00)}\\         
9     &98.57(4.52) &99.29(2.26) &95.00(7.57) &\textbf{100.00(0.00)}\\         
10     &88.29(1.93) &88.22(1.99) &68.98(10.02) &\textbf{93.22(1.33)}\\         
11     &83.63(2.20) &82.46(2.29) &63.02(3.59) &\textbf{87.87(1.38)}\\         
12     &89.72(1.27) &89.47(2.13) &78.36(4.49) &\textbf{94.55(2.37)}\\         
13     &\textbf{100.00(0.00)} &99.93(0.22) &98.41(2.28) &\textbf{100.00(0.00)}\\         
14     &96.45(0.80) &96.39(1.06) &84.58(4.79) &\textbf{97.26(0.77)}\\         
15     &98.65(1.71) &96.90(3.52) &90.39(5.48) &\textbf{99.74(0.62)}\\         
16     &\textbf{96.82(1.81)} &93.79(2.62) &89.24(3.88) &95.30(1.51)\\         
\hline
OA &90.39(0.70) &89.36(1.36) &72.10(3.63) &\textbf{93.22(0.49)}\\         
AA &94.84(0.40) &94.19(0.93) &82.80(3.78) &\textbf{96.44(0.21)}\\         
Kappa&88.94(0.81) &87.75(1.56) &68.31(4.05) &\textbf{92.17(0.56)}\\         
\hline
\end{tabular}
\label{tab:tab1}
\end{table}
\par Experiments on KSC datasets are carried out to further validate the proposed method. The same experimental settings are retained and the detailed classification results can be seen in Table \uppercase\expandafter{\romannumeral2}. The optimal results under each class and criterion are shown in bold among the table.
\begin{table}[!h]
\centering  
\caption{Classification Results on KSC Dataset. Numbers in the Parenthesis is the Standard Variances of the Accuracies Obtained in Repeated Experiments.}
\renewcommand\arraystretch{1.25}
\begin{tabular}{C{0.5cm}C{1.5cm}C{1.55cm}C{1.62cm}C{1.5cm}}  
\hline
Class    &CM &SE+CM \cite{SENET} &BW+CM \cite{BandWeighting} &BAM+CM \\
\hline
1   &99.06(0.72)    &\textbf{99.21(0.77)}    &92.22(4.34)    &99.00(1.12)\\
2   &66.67(6.41)    &72.79(10.9)    &52.79(8.24)    &\textbf{86.80(6.23)}\\
3   &91.99(3.20)     &91.56(3.81)    &81.52(6.35)    &\textbf{93.64(3.59)}\\
4   &66.43(6.32)    &67.18(8.15)    &57.40(4.95)     &\textbf{78.59(5.58)}\\
5   &63.45(4.50)     &63.52(8.21)    &65.66(3.33)    &\textbf{71.24(6.44)}\\
6   &72.51(3.67)    &73.48(3.03)    &41.11(5.44)    &\textbf{76.38(4.38)}\\
7   &96.11(3.40)     &\textbf{97.58(3.14)}    &92.53(5.31)    &94.00(4.79)\\
8   &84.93(7.46)    &\textbf{96.77(2.54)}    &70.41(6.55)    &95.84(1.91)\\
9   &96.65(2.22)    &98.42(1.36)    &93.85(1.04)    &\textbf{99.79(0.46)}\\
10  &97.06(2.02)    &97.50(1.50)      &89.62(3.02)    &\textbf{99.01(1.01)}\\
11  &97.70(1.59)    &97.62(1.63)    &84.52(2.23)    &\textbf{98.04(1.78)}\\
12  &96.58(1.79)    &98.32(1.62)    &96.09(0.56)    &\textbf{99.21(0.93)}\\
13  &\textbf{100.00(0.00)}    &\textbf{100.00(0.00)}   &99.27(0.94)    &\textbf{100.00(0.00)} \\
\hline
OA  &91.38(0.95)&93.12(0.91)&84.19(0.88)&\textbf{95.06(0.51)}\\
AA  &86.86(1.31)&88.76(1.53)&78.23(1.39)&\textbf{91.66(0.95)}\\
Kappa &90.40(1.06)&92.34(1.01)&82.36(0.99)&\textbf{94.50(0.57)}\\
\hline
\end{tabular}
\label{tab:tab2}
\end{table}
\par Through the analysis of Table \uppercase\expandafter{\romannumeral2}, we can see that the trend of comparison between the accuracy of each method is similar to the former. The proposed method still achieves the best classification results on KSC dataset. Compared with the CM, the proposed method achieves $4\%$-$5\%$ improvement under involved criteria. Unlike before, although SE block does not consider the application background of band selection, its addition has also achieved a slight performance improvement. However, the effect of SE block is still at least $2\%$ less than it of BAM.
\par Based on the experimental results of the above two parts, we can see that the proposed BAM can be directly added to a CM without deliberate design to improve its performance. Compared with the existing modules with similar roles, BAM can improve the accuracy more greatly. Thus, the proposed BAM can be regarded as a plug-and-play supplementary component to most of the mainstream CNNs in HSIs classification.

\subsection{Effect of hyperparameters}
In this section, we present a detailed analysis and evaluation of the influence of hyperparameters on the performance of the proposed method. In the analysis, Indian Pines dataset and the VGGNet based CM are used to explore the changing trend of the classification accuracy. The experimental results are shown in Fig. 4.
\begin{figure}[!h]
\begin{minipage}{0.49\linewidth}
  \centerline{\includegraphics[width=4.5cm,height=3.0cm]{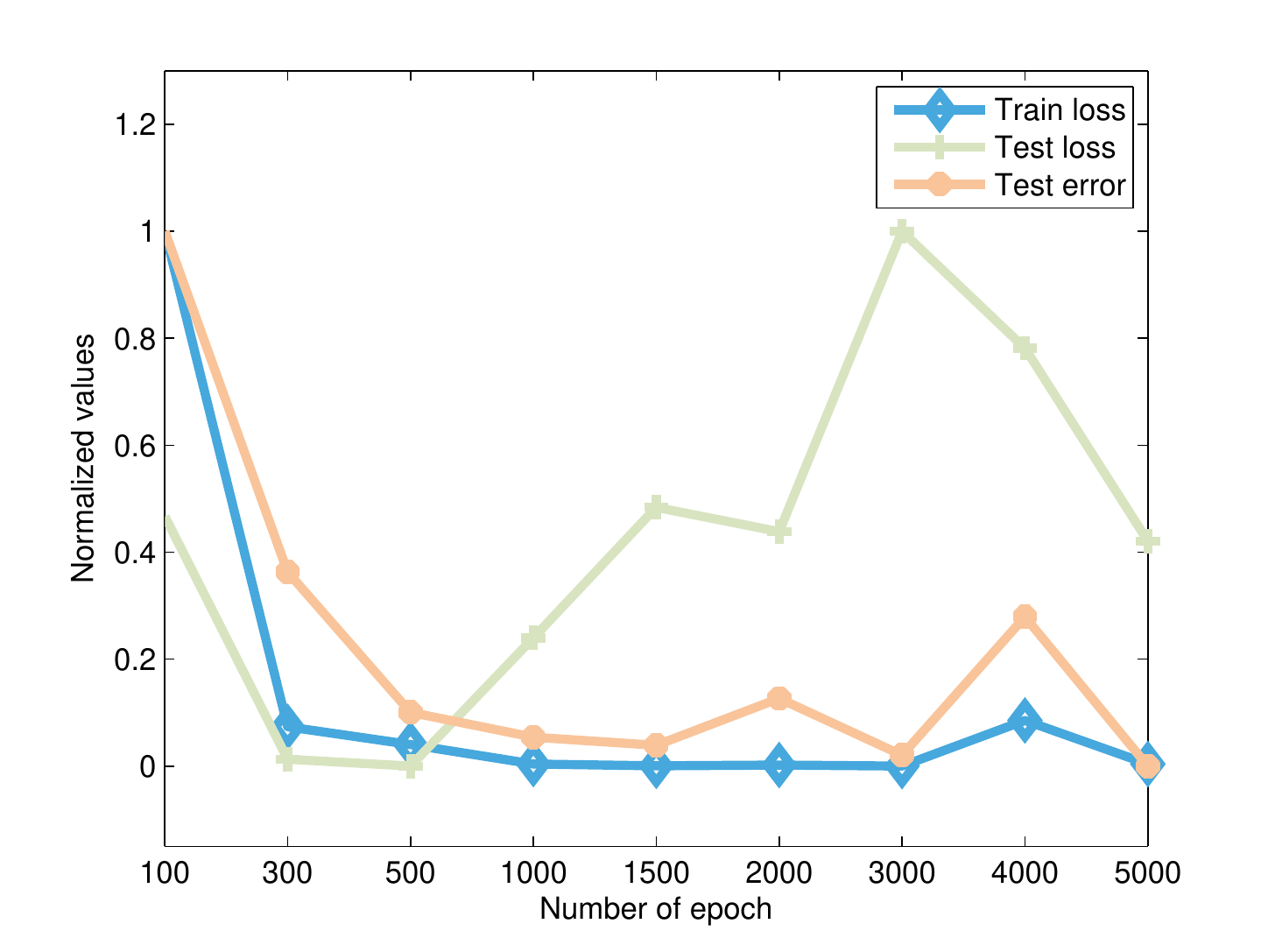}}
  \centerline{(a)}
\end{minipage}
\hfill
\begin{minipage}{0.49\linewidth}
  \centerline{\includegraphics[width=4.50cm,height=3.0cm]{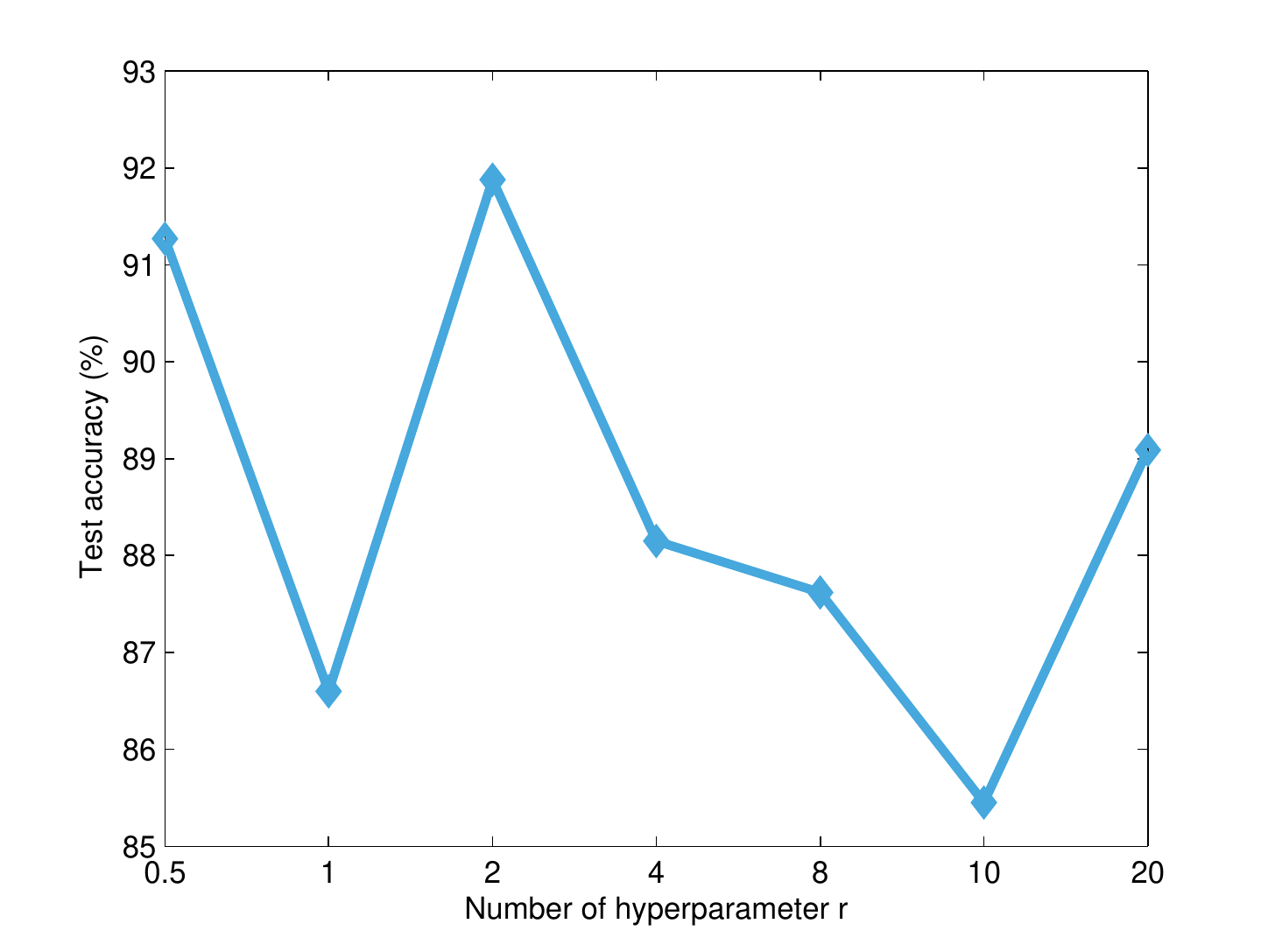}}
  \centerline{(b)}
\end{minipage}
\caption{Effect of two hyperparameters epoch and $r$ (ratio of information aggregation in 1D convolution layers) of the proposed method on Indian Pines dataset.}
\label{fig4}
\end{figure}
\par Fig. 4(a)-(b) shows the OA of the proposed method under different value of epoch and $r$. From Fig. 4(a), it can be seen that the loss of the proposed model changes slightly when the epoch is greater than 500, which means that the training process is close to convergence. Besides, too many training times lead to the increase of the loss on both training and testing set, and reduce the testing accuracy to a certain extent. Further, we can see from Fig. 4(b) that hyperparameter $r=0.5$ and $r=2$ obtain better classification results. Based on the above empirical knowledge, the training set is reused 1000 times and the information compression ratio $r$ is set to 2 to achieve better performance in the experiments.
\subsection{Combination with advanced CNNs}
In this section, we test the performance of some advanced CNNs combined with the BAM in order to further verify the general applicability of the proposed BAM, and also to obtain the better results of HSIs classification. Four advanced backbones including two-stream CNN (TSCNN) \cite{twostream,Yang2016}, the network with depthwise separable convolutions (Xception) \cite{xception}, the network with residual connection (ResNet) \cite{ResNet} and the network with densely connection (DenseNet) \cite{DenseNet} are chosen for testing.
\begin{figure}[!h]
\begin{minipage}{0.49\linewidth}
  \centerline{\includegraphics[width=5.0cm,height=3.0cm]{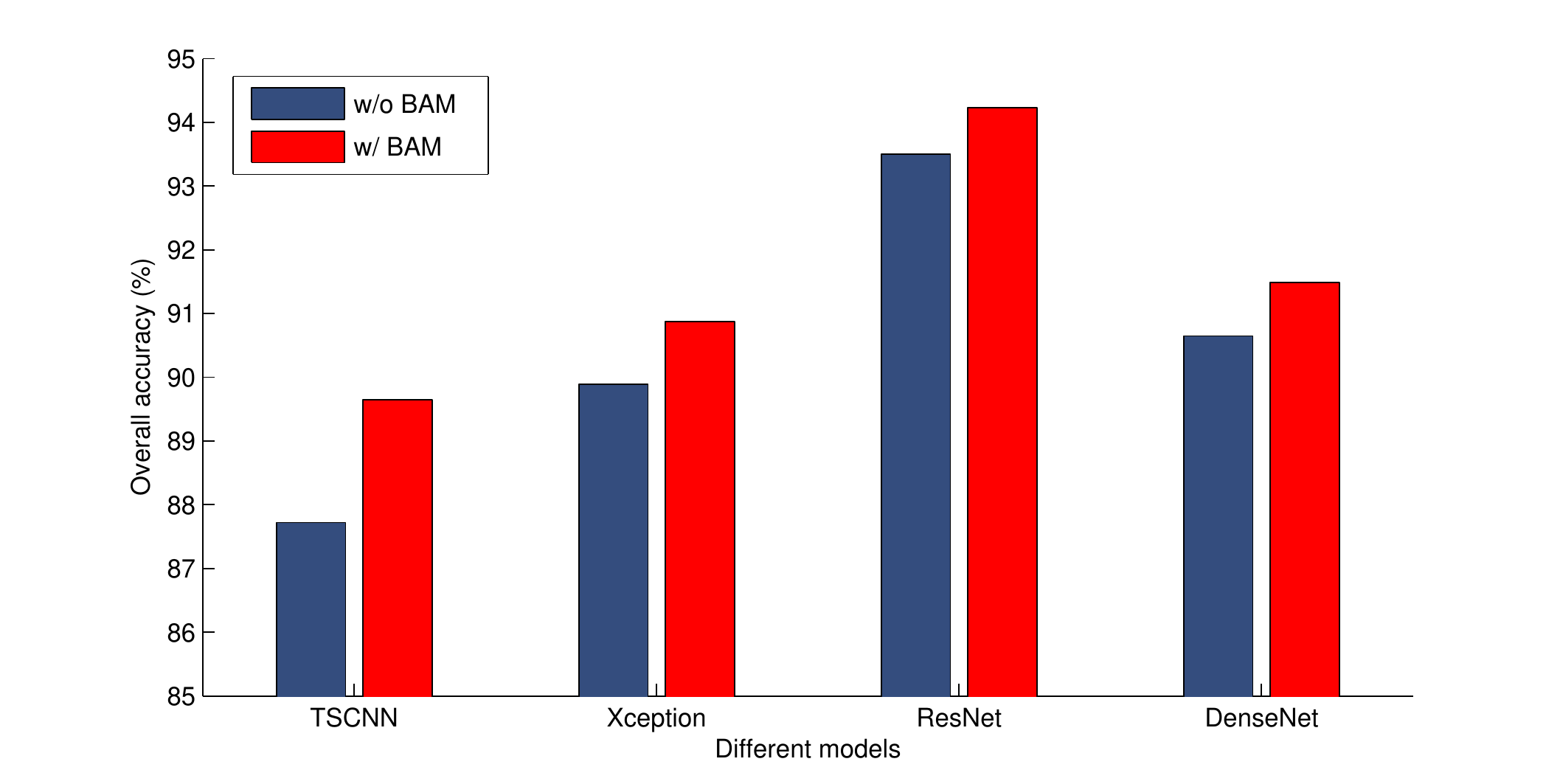}}
  \centerline{(a)}
\end{minipage}
\hfill
\begin{minipage}{0.49\linewidth}
  \centerline{\includegraphics[width=5.0cm,height=3.0cm]{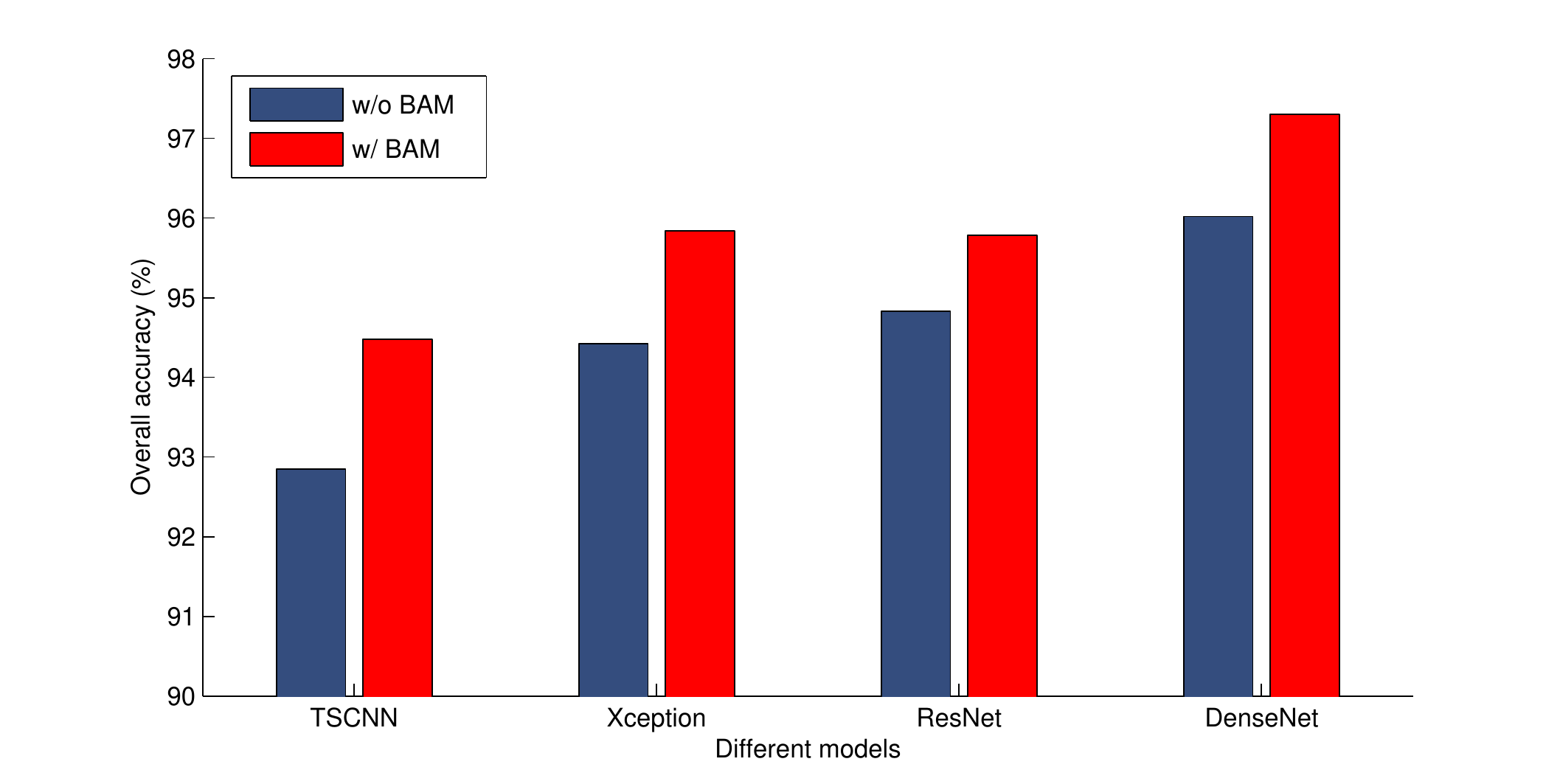}}
  \centerline{(b)}
\end{minipage}
\caption{Experimental results of different CNN backbones combined with the proposed BAM on Indian Pines and KSC datasets.}
\label{fig5}
\end{figure}
\par As shown in Fig. 7, the backbone with the BAM achieves higher classification accuracy in both datasets, which proves that the BAM has wider applicability and can be used as a plug-and-play module to improve the performance of most models for HSI classification.

\section{Conclusion}\label{sec:4}
In this letter, we propose a novel deep learning based HSIs classification framework which fully considers the redundancy and noise in the band of HSIs. A well-designed BAM is embedded in the ordinary CNN to implement an end-to-end network with the capability of band selection. This module absorbs the experience of visual attention mechanism so it can adaptively stimulate the bands which are beneficial to the improvement of classification accuracy, while suppressing the invalid bands. The proposed band attention based deep classification framework has better adaptability to the task of HSIs classification than the mainstream CNNs since it is customized according to the characteristics of HSIs. Abundant numerical experiments not only reveal the influence of hyperparameters on classification accuracy, but also show that the proposed BAM can be a plug-and-play module to improve the accuracy of CNNs in HSIs classification.


%

%
%
%
%
%
%
%

\bibliographystyle{IEEEtran}
\bibliography{IEEEabrv,refference}
\end{document}